# Artificial Intelligence and Data Science in the Automotive Industry

Martin Hofmann[1], Florian Neukart[2,3], Thomas Bäck[3]

[*1,2]Volkswagen AG, [3]Leiden University

[*1]martin.hofmann@volkswagen.de; [*2]florian.neukart@vw.com; [*3]t.h.w.baeck@liacs.leidenuniv.nl

**Abstract**

Data science and machine learning are the key technologies when it comes to the processes and products with automatic learning and optimization to be used in the automotive industry of the future. This article defines the terms "data science" (also referred to as "data analytics") and "machine learning" and how they are related. In addition, it defines the term "*optimizing analytics*" and illustrates the role of automatic optimization as a key technology in combination with data analytics. It also uses examples to explain the way that these technologies are currently being used in the automotive industry on the basis of the major subprocesses in the automotive value chain (development, procurement; logistics, production, marketing, sales and after-sales, connected customer). Since the industry is just starting to explore the broad range of potential uses for these technologies, visionary application examples are used to illustrate the revolutionary possibilities that they offer. Finally, the article demonstrates how these technologies can make the automotive industry more efficient and enhance its customer focus throughout all its operations and activities, extending from the product and its development process to the customers and their connection to the product.

## 1 Introduction

Data science and machine learning are now key technologies in our everyday lives, as we can see in a multitude of applications, such as voice recognition in vehicles and on cell phones, automatic facial and traffic sign recognition, as well as chess and, more recently, Go machine algorithms[1] which humans can no longer beat. The analysis of large data volumes based on search, pattern recognition, and learning algorithms provides insights into the behavior of processes, systems, nature, and ultimately people, opening the door to a world of fundamentally new possibilities. In fact, the now already implementable idea of autonomous driving is virtually a tangible reality for many drivers today with the help of lane keeping assistance and adaptive cruise control systems in the vehicle.

The fact that this is just the tip of the iceberg, even in the automotive industry, becomes readily apparent when one considers that, at the end of 2015, Toyota and Tesla's founder, Elon Musk, each announced investments amounting to one billion US dollars in artificial intelligence research and development almost at the same time. The trend towards connected, autonomous, and artificially intelligent systems that continuously learn from data and are able to make optimal decisions is advancing in ways that are simply revolutionary, not to mention fundamentally important to many industries. This includes the automotive industry, one of the key industries in Germany, in which international competitiveness will be influenced by a new factor in the near future – namely the new technical and service offerings that can be provided with the help of data science and machine learning.

This article provides an overview of the corresponding methods and some current application examples in the automotive industry. It also outlines the potential applications to be expected in this industry very soon. Accordingly, sections 2 and 3 begin by addressing the subdomains of data mining (also referred to as "big data analytics") and artificial intelligence, briefly summarizing the corresponding processes, methods, and areas of application and presenting them in context. Section 4 then provides an overview of current application examples in the automotive industry based on the stages in the industry's value chain –from development to production and logistics through to the end customer. Based on such an example, section 5 describes the vision for future applications using three examples: one in which vehicles play the role of autonomous agents that interact with each other in cities, one that covers integrated production optimization, and one that describes companies themselves as autonomous agents.

---

[1] D. Silver et. al.: Mastering the Game of Go with Deep Neural Networks and Tree Search, Nature 529, 484-489 (January 28, 2016).

Whether these visions will become a reality in this or any other way cannot be said with certainty at present – however, we can safely predict that the rapid rate of development in this area will lead to the creation of completely new products, processes, and services, many of which we can only imagine today. This is one of the conclusions drawn in section 6, together with an outlook regarding the potential future effects of the rapid rate of development in this area.

## 2 The data mining process

Gartner uses the term "*prescriptive analytics*" to describe the highest level of ability to make business decisions on the basis of data-based analyses. This is illustrated by the question "what should I do?" and *prescriptive analytics* supplies the required decision-making support, if a person is still involved, or automation if this is no longer the case.

The levels below this, in ascending order in terms of the use and usefulness of AI and data science, are defined as follows: *descriptive analytics* ("what has happened?"), *diagnostic analytics* ("why did it happen?"), and *predictive analytics* ("what will happen?") (see Figure 1). The last two levels are based on data science technologies, including data mining and statistics, while *descriptive analytics* essentially uses traditional business intelligence concepts (data warehouse, OLAP).

In this article, we seek to replace the term "*prescriptive analytics*" with the term "*optimizing analytics*." The reason for this is that a technology can "prescribe" many things, while, in terms of implementation within a company, the goal is always to make something "better" with regard to target criteria or quality criteria. This optimization can be supported by search algorithms, such as evolutionary algorithms in nonlinear cases and operation research (OR) methods in – much rarer – linear cases. It can also be supported by application experts who take the results from the data mining process and use them to draw conclusions regarding process improvement. One good example are the decision trees learned from data, which application experts can understand, reconcile with their own expert knowledge, and then implement in an appropriate manner. Here too, the application is used for *optimizing* purposes, admittedly with an intermediate human step.

Within this context, another important aspect is the fact that multiple criteria required for the relevant application often need to be optimized at the same time, meaning that multi-criteria optimization methods – or, more generally, multi-criteria decision-making support methods – are necessary. These methods can then be used in order to find the best possible compromises between conflicting goals. The examples mentioned include the frequently occurring conflicts between cost and quality, risk and profit, and, in a more technical example, between the weight and passive occupant safety of a body.

| Optimzing Analytics | "What am I supposed to do?" | Decision support, multicriterial optimization |
|---|---|---|
| Predictive Analytics | "What will happen?" | Modelling |
| Diagnostic Analytics | "Why did it happen?" | BI, Modelling |
| Descriptive Analytics | "What happened?" | Business Intelligence |

**Figure 1: The four levels of data analysis usage within a company**

These four levels form a framework, within which it is possible to categorize data analysis competence and potential benefits for a company in general. This framework is depicted in Figure 1 and shows the four layers which build upon each other, together with the respective technology category required for implementation.

The traditional *Cross-Industry Standard Process for Data Mining* (CRISP-DM)[2] includes no optimization or decision-making support whatsoever. Instead, based on the *business understanding*, *data understanding*, *data preparation*, *modeling*, and *evaluation* sub-steps, CRISP proceeds directly to the *deployment* of results in business processes. Here too, we propose an additional *optimization* step that in turn comprises multi-criteria optimization and decision-making support. This approach is depicted schematically in Figure 2.

---
2

https://en.wikipedia.org/wiki/Cross_Industry_Standard_Process_for_Data_Mining

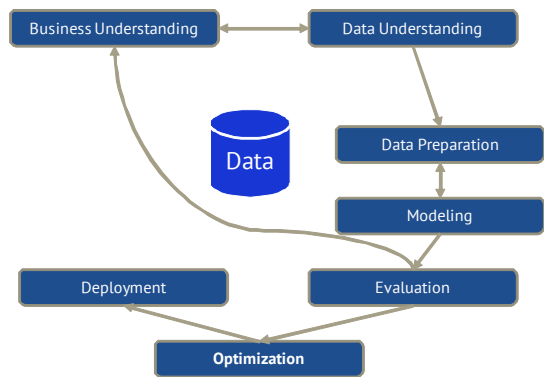

**Figure 2: Traditional CRISP-DM process with an additional optimization step**

It is important to note that the original CRISP model deals with a largely iterative approach used by data scientists to analyze data manually, which is reflected in the iterations between *business understanding* and *data understanding* as well as *data preparation* and *modeling*. However, evaluating the modeling results with the relevant application experts in the *evaluation* step can also result in having to start the process all over again from the *business understanding* sub-step, making it necessary to go through all the sub-steps again partially or completely (e.g., if additional data needs to be incorporated).

The manual, iterative procedure is also due to the fact that the basic idea behind this approach – as up-to-date as it may be for the majority of applications – is now almost 20 years old and certainly only partially compatible with a big data strategy. The fact is that, in addition to the use of *nonlinear modeling methods* (in contrast to the usual generalized linear models derived from statistical modeling) and *knowledge extraction* from data, data mining rests on the fundamental idea that *models can be derived from data with the help of algorithms and that this modeling process can run automatically for the most part* – because the algorithm "does the work."

In applications where a large number of models need to be created, for example for use in making forecasts (e.g., sales forecasts for individual vehicle models and markets based on historical data), automatic modeling plays an important role. The same applies to the use of online data mining, in which, for example, forecast models (e.g., for forecasting product quality) are not only constantly used for a production process, but also adapted (i.e., retrained) continuously whenever individual process aspects change (e.g., when a new raw material batch is used). This type of application requires the technical ability to automatically generate data, and integrate and process it in such a way that data mining algorithms can be applied to it. In addition, automatic modeling and automatic optimization are necessary in order to update models and use them as a basis for generating optimal proposed actions in online applications. These actions can then be communicated to the process expert as a suggestion or – especially in the case of continuous production processes – be used directly to control the respective process. If sensor systems are also integrated directly into the production process – to collect data in real time – this results in a self-learning cyber-physical system [3] that facilitates implementation of the Industry 4.0[4] vision in the field of production engineering.

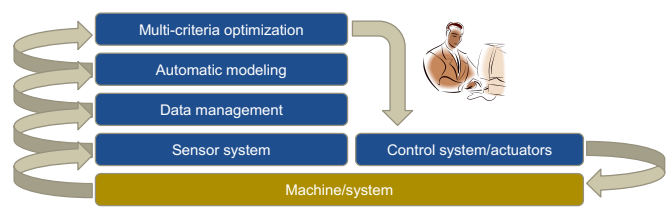

**Figure 3: Architecture of an Industry 4.0 model for optimizing analytics**

This approach is depicted schematically in Figure 3. Data from the system is acquired with the help of sensors and integrated into the data management system. Using this as a basis, forecast models for the system's relevant outputs (quality, deviation from target value, process variance, etc.) are used continuously in order to forecast the system's output. Other machine learning options can be used within this context in order, for example, to predict maintenance results (predictive maintenance) or to identify anomalies in the process. The corresponding models are monitored continuously and, if necessary, automatically retrained if any process drift is observed. Finally, the multi-criteria optimization uses the models to continuously compute optimum setpoints for the system control. Human process experts can also be integrated here by using the system as a suggestion generator so that a process expert can evaluate the generated suggestions before they are implemented in the original system.

---

[3] Systems "in which information and software components are connected to mechanical and electronic components and in which data is transferred and exchanged, and monitoring and control tasks are carried out, in real-time using infrastructures such as the Internet." (Translation of the following article in Gabler Wirtschaftslexikon, Springer: http://wirtschaftslexikon.gabler.de/Definition/cyber-physische-systeme.html).

[4] Industry 4.0 is defined therein as "a marketing term that is also used in science communication and refers to a 'future project' of the German federal government. The so-called 'Fourth Industrial Revolution' is characterized by the customization and hybridization of products and the integration of customers and business partners into business processes." (Translation of the following article in Gabler Wirtschaftslexikon, Springer: http://wirtschaftslexikon.gabler.de/Definition/industrie-4-0.html).

In order to differentiate it from "traditional" data mining, the term "big data" is frequently defined now with three (sometimes even four or five) essential characteristics: *volume*, *velocity*, and *variety*, which refer to the large volume of data, the speed at which data is generated, and the heterogeneity of the data to be analyzed, which can no longer be categorized into the conventional relational database schema. *Veracity*, i.e., the fact that large uncertainties may also be hidden in the data (e.g., measurement inaccuracies), and finally *value*, i.e., the value that the data and its analysis represents for a company's business processes, are often cited as additional characteristics. So it is not just the pure data volume that distinguishes previous data analytics methods from big data, but also other technical factors that require the use of new methods– such as Hadoop and MapReduce – with appropriately adapted data analysis algorithms in order to allow the data to be saved and processed. In addition, so-called "in-memory databases" now also make it possible to apply traditional learning and modeling algorithms in main memory to large data volumes.

This means that if one were to establish a hierarchy of data analysis and modeling methods and techniques, then, in very simplistic terms, statistics would be a subset of data mining, which in turn would be a subset of big data. Not every application requires the use of data mining or big data technologies. However, a clear trend can be observed, which indicates that the necessities and possibilities involved in the use of data mining and big data are growing at a very rapid pace as increasingly large data volumes are being collected and linked across all processes and departments of a company. Nevertheless, conventional hardware architecture with additional main memory is often more than sufficient for analyzing large data volumes in the gigabyte range.

Although *optimizing analytics* is of tremendous importance, it is also crucial to always be open to the broad variety of applications when using artificial intelligence and machine learning algorithms. The wide range of learning and search methods, with potential use in applications such as image and language recognition, knowledge learning, control and planning in areas such as production and logistics, among many others, can only be touched upon within the scope of this article.

## 3. The pillars of artificial intelligence

An early definition of artificial intelligence from the IEEE Neural Networks Council was "the study of how to make computers do things at which, at the moment, people are better."[5] Although this still applies, current research is also focused on improving the way that software does things at which computers have always been better, such as analyzing large amounts of data. Data is also the basis for developing artificially intelligent software systems not only to collect information, but also to:

- Learn
- Understand and interpret information
- Behave adaptively
- Plan
- Make inferences
- Solve problems
- Think abstractly
- Understand and interpret ideas and language

### 3.1 Machine learning

At the most general level, machine learning (ML) algorithms can be subdivided into two categories: supervised and unsupervised, depending on whether or not the respective algorithm requires a target variable to be specified.

*Supervised learning algorithms*

Apart from the input variables (predictors), supervised learning algorithms also require the known target values (labels) for a problem. In order to train an ML model to identify traffic signs using cameras, images of traffic signs – preferably with a variety of configurations – are required as input variables. In this case, light conditions, angles, soiling, etc. are compiled as noise or blurring in the data; nonetheless, it must be possible to recognize a traffic sign in rainy conditions with the same accuracy as when the sun is shining. The labels, i.e., the correct designations, for such data are normally assigned manually. This correct set of input variables and their correct classification constitute a training data set. Although we only have one image per training data set in this case, we still speak of multiple input variables, since ML algorithms find relevant features in training data and learn how these features and the class assignment for the classification task indicated in the example are associated. Supervised learning is used primarily to predict numerical values (regression) and for classification purposes (predicting the appropriate class), and the corresponding data is not limited to a specific format – ML algorithms are more than capable of processing images, audio files, videos, numerical data, and text. Classification examples include object recognition (traffic signs, objects in front of a vehicle, etc.), face recognition, credit risk assessment, voice recognition, and customer churn, to name but a few.

---

[5] E. Rich, K. Knight: Artificial Intelligence, 5, 1990

Regression examples include determining continuous numerical values on the basis of multiple (sometimes hundreds or thousands) input variables, such as a self-driving car calculating its ideal speed on the basis of road and ambient conditions, determining a financial indicator such as gross domestic product based on a changing number of input variables (use of arable land, population education levels, industrial production, etc.), and determining potential market shares with the introduction of new models. Each of these problems is highly complex and cannot be represented by simple, linear relationships in simple equations. Or, to put it another way that more accurately represents the enormous challenge involved: the necessary expertise does not even exist.

*Unsupervised learning algorithms*

Unsupervised learning algorithms do not focus on individual target variables, but instead have the goal of characterizing a data set in general. Unsupervised ML algorithms are often used to group (cluster) data sets, i.e., to identify relationships between individual data points (that can consist of any number of attributes) and group them into clusters. In certain cases, the output from unsupervised ML algorithms can in turn be used as an input for supervised methods. Examples of unsupervised learning include forming customer groups based on their buying behavior or demographic data, or clustering time series in order to group millions of time series from sensors into groups that were previously not obvious.

In other words, machine learning is the area of artificial intelligence (AI) that enables computers to learn without being programmed explicitly. Machine learning focuses on developing programs that grow and change by themselves as soon as new data is provided. Accordingly, processes that can be represented in a flowchart are not suitable candidates for machine learning – in contrast, everything that requires dynamic and changing solution strategies and cannot be constrained to static rules is potentially suitable for solution with ML. For example, ML is used when:

- No relevant human expertise exists
- People are unable to express their expertise
- The solution changes over time
- The solution needs to be adapted to specific cases

In contrast to statistics, which follows the approach of making inferences based on samples, computer science is interested in developing efficient algorithms for solving optimization problems, as well as in developing a representation of the model for evaluating inferences. Methods frequently used for optimization in this context include so-called "evolutionary algorithms" (genetic algorithms, evolution strategies), the basic principles of which emulate natural evolution[6]. These methods are very efficient when applied to complex, nonlinear optimization problems.

Even though ML is used in certain data mining applications, and both look for patterns in data, ML and data mining are not the same thing. Instead of extracting data that people can understand, as is the case with data mining, ML methods are used by programs to improve their own understanding of the data provided. Software that implements ML methods recognizes patterns in data and can dynamically adjust the behavior based on them. If, for example, a self-driving car (or the software that interprets the visual signal from the corresponding camera) has been trained to initiate a braking maneuver if a pedestrian appears in front it, this must work with all pedestrians regardless of whether they are short, tall, fat, thin, clothed, coming from the left, coming from the right, etc. In turn, the vehicle must not brake if there is a stationary garbage bin on the side of the road.

The level of complexity in the real world is often greater than the level of complexity of an ML model, which is why, in most cases, an attempt is made to subdivide problems into subproblems and then apply ML models to these subproblems. The output from these models is then integrated in order to permit complex tasks, such as autonomous vehicle operation, in structured and unstructured environments.

---

[6] Th. Bäck, D.B. Fogel, Z. Michalewicz: Handbook of Evolutionary Computation, Institute of Physics Publishing, New York, 1997.

## 3.2 Computer vision

Computer vision (CV) is a very wide field of research that merges scientific theories from various fields (as is often the case with AI), starting from biology, neuroscience, and psychology and extending all the way to computer science, mathematics, and physics. First, it is important to know how an image is produced physically. Before light hits sensors in a two-dimensional array, it is refracted, absorbed, scattered, or reflected, and an image is produced by measuring the intensity of the light beams through each element in the image (pixel). The three primary focuses of CV are:

- Reconstructing a scene and the point from which the scene is observed based on an image, an image sequence, or a video.
- Emulating biological visual perception in order to better understand which physical and biological processes are involved, how the wetware works, and how the corresponding interpretation and understanding work.
- Technical research and development focuses on efficient, algorithmic solutions – when it comes to CV software, problem-specific solutions that only have limited commonalities with the visual perception of biological organisms are often developed.

All three areas overlap and influence each other. If, for example, the focus in an application is on obstacle recognition in order to initiate an automated braking maneuver in the event of a pedestrian appearing in front of the vehicle, the most important thing is to identify the pedestrian as an obstacle. Interpreting the entire scene – e.g., understanding that the vehicle is moving towards a family having a picnic in a field – is not necessary in this case. In contrast, understanding a scene is an essential prerequisite if context is a relevant input, such as is the case when developing domestic robots that need to understand that an occupant who is lying on the floor not only represents an obstacle that needs to be evaded, but is also probably not sleeping and a medical emergency is occurring.

Vision in biological organisms is regarded as an active process that includes controlling the sensor and is tightly linked to successful performance of an action [7]. Consequently,[8] CV systems are not passive either. In other words, the system must:

- Be continuously provided with data via sensors (streaming)
- Act based on this data stream

Having said that, the goal of CV systems is not to understand scenes in images – first and foremost, the systems must extract the relevant information for a specific task from the scene. This means that they must identify a "region of interest" that will be used for processing. Moreover, these systems must feature short response times, since it is probable that scenes will change over time and that a heavily delayed action will not achieve the desired effect. Many different methods have been proposed for object recognition purposes ("what" is located "where" in a scene), including:

- Object detectors, in which case a window moves over the image and a filter response is determined for each position by comparing a template and the sub-image (window content), with each new object parameterization requiring a separate scan. More sophisticated algorithms simultaneously make calculations based on various scales and apply filters that have been learned from a large number of images.
- Segment-based techniques extract a geometrical description of an object by grouping pixels that define the dimensions of an object in an image. Based on this, a fixed feature set is computed, i.e., the features in the set retain the same values even when subjected to various image transformations, such as changes in light conditions, scaling, or rotation. These features are used to clearly identify objects or object classes, one example being the aforementioned identification of traffic signs.

---

[7] R. Bajcsy: Active perception, Proceedings of the IEEE, 76:996-1005, 1988
[8] J. L. Crowley, H. I. Christensen: Vision as a Process: Basic Research on Computer Vision Systems, Berlin: Springer, 1995

- Alignment-based methods use parametric object models that are trained on data[9,10]. Algorithms search for parameters, such as scaling, translation, or rotation, that adapt a model optimally to the corresponding features in the image, whereby an approximated solution can be found by means of a reciprocal process, i.e., by features, such as contours, corners, or others, "selecting" characteristic points in the image for parameter solutions that are compatible with the found feature.

With object recognition, it is necessary to decide whether algorithms need to process 2-D or 3-D representations of objects – 2-D representations are very frequently a good compromise between accuracy and availability. Current research (deep learning) shows that even distances between two points based on two 2-D images captured from different points can be accurately determined as an input. In daylight conditions and with reasonably good visibility, this input can be used in addition to data acquired with laser and radar equipment in order to increase accuracy – moreover, a single camera is sufficient to generate the required data. In contrast to 3-D objects, no shape, depth, or orientation information is directly encoded in 2-D images. Depth can be encoded in a variety of ways, such as with the use of laser or stereo cameras (emulating human vision) and structured light approaches (such as Kinect). At present, the most intensively pursued research direction involves the use of superquadrics – geometric shapes defined with formulas, which use any number of exponents to identify structures such as cylinders, cubes, and cones with round or sharp edges. This allows a large variety of different basic shapes to be described with a small set of parameters. If 3-D images are acquired using stereo cameras, statistical methods (such as generating a stereo point cloud) are used instead of the aforementioned shape-based methods, because the data quality achieved with stereo cameras is poorer than that achieved with laser scans.

Other research directions include tracking[11,12], contextual scene understanding,[13,14] and monitoring[15], although these aspects are currently of secondary importance to the automotive industry.

### 3.3 Inference and decision-making

This field of research, referred to in the literature as "knowledge representation & reasoning" (KRR), focuses on designing and developing data structures and inference algorithms. Problems solved by making inferences are very often found in applications that require interaction with the physical world (humans, for example), such as generating diagnostics, planning, processing natural languages, answering questions, etc. KRR forms the basis for AI at the human level.

Making inferences is the area of KRR in which data-based answers need to be found without human intervention or assistance, and for which data is normally presented in a formal system with distinct and clear semantics. Since 1980, it has been assumed that the data involved is a mixture of simple and complex structures, with the former having a low degree of computational complexity and forming the basis for research involving large databases. The latter are presented in a language with more expressive power, which requires less space for representation, and they correspond to generalizations and fine-grained information.

Decision-making is a type of inference that revolves primarily around answering questions regarding preferences between activities, for example when an autonomous agent attempts to fulfill a task for a person. Such decisions are very frequently made in a dynamic domain which changes over the course of time and when actions are executed. An example of this is a self-driving car that needs to react to changes in traffic.

---

*Logic and combinatorics*

Mathematical logic is the formal basis for many applications in the real world, including calculation theory, our legal system and corresponding arguments, and theoretical developments and evidence in the field of research and development. The initial vision was to represent every type of knowledge in the form of logic and use universal algorithms to make inferences from it, but a number of challenges arose – for example, not all types of knowledge can be represented simply. Moreover, compiling the knowledge required for complex applications can become very complex, and it is not easy to learn this type of knowledge in a logical, highly expressive language.[16] In addition, it is not easy to make inferences with the required highly expressive language – in extreme cases, such scenarios cannot be implemented computationally, even if the first two challenges are overcome. Currently, there are three ongoing debates on this subject, with the first one focusing on the argument that logic is unable to represent many concepts, such as space, analogy, shape, uncertainty, etc., and consequently cannot be included as an active part in developing AI to a human level. The counterargument states that logic is simply one of many tools. At present, the combination of representative expressiveness, flexibility, and clarity cannot be achieved with any other method or system. The second debate revolves around the argument that logic is too slow for making inferences and will therefore never play a role in a productive system. The counterargument here is that ways exist to approximate the inference process with logic, so processing is drawing close to remaining within the required time limits, and progress is being made with regard to logical inference. Finally, the third debate revolves around the argument that it is extremely difficult, or even impossible, to develop systems based on logical axioms into applications for the real world. The counterarguments in this debate are primarily based on the research of individuals currently researching techniques for learning logical axioms from natural-language texts.

In principle, a distinction is made between four different types of logic[17] which are not discussed any further in this article:

- Propositional logic
- First-order predicate logic
- Modal logic
- Non-monotonic logic

Automated decision-making, such as that found in autonomous robots (vehicles), WWW agents, and communications agents, is also worth mentioning at this point. This type of decision-making is particularly relevant when it comes to representing expert decision-making processes with logic and automating them. Very frequently, this type of decision-making process takes account of the dynamics of the surroundings, for example when a transport robot in a production plant needs to evade another transport robot. However, this is not a basic prerequisite, for example, if a decision-making process without a clearly defined direction is undertaken in future, e.g., the decision to rent a warehouse at a specific price at a specific location. Decision-making as a field of research encompasses multiple domains, such as computer science, psychology, economics, and all engineering disciplines. Several fundamental questions need to be answered to enable development of automated decision-making systems:

- Is the domain dynamic to the extent that a sequence of decisions is required or static in the sense that a single decision or multiple simultaneous decisions need to be made?
- Is the domain deterministic, non-deterministic, or stochastic?
- Is the objective to optimize benefits or to achieve a goal?
- Is the domain known to its full extent at all times? Or is it only partially known?

Logical decision-making problems are non-stochastic in nature as far as planning and conflicting behavior are concerned. Both require that the available information regarding the initial and intermediate states be complete, that actions have exclusively deterministic, known effects, and that a specific defined goal exists. These problem types are often applied in the real world, for example in robot control, logistics, complex behavior in the WWW, and in computer and network security.

In general, planning problems consist of an initial (known)

---

[16] N. Lavarac, S. Dzeroski: Inductive Logic Programming, Vol. 3: Non-Monotonic Reasoning and Uncertain Reasoning, Oxford University Press: Oxford, 1994

[17] K. Frankish, W. M. Ramsey: The Cambridge Handbook of Artificial Intelligence, Cambridge: Cambridge University Press, 2014

situation, a defined goal, and a set of permitted actions or transitions between steps. The result of a planning process is a sequence or set of actions that, when executed correctly, change the executing entity from an initial state to a state that meets the target conditions. Computationally speaking, planning is a difficult problem, even if simple problem specification languages are used. Even when relatively simple problems are involved, the search for a plan cannot run through all state-space representations, as these are exponentially large in the number of states that define the domains. Consequently, the aim is to develop efficient algorithms that represent sub-representations in order to search through these with the hope of achieving the relevant goal. Current research is focused on developing new search methods and new representations for actions and states, which will make planning easier. Particularly when one or more agents acting against each other are taken into account, it is crucial to find a balance between learning and decision-making – exploration for the sake of learning while decisions are being made can lead to undesirable results.

Many problems in the real world are problems with dynamics of a stochastic nature. One example of this is buying a vehicle with features that affect its value, of which we are unaware. These dependencies influence the buying decision, so it is necessary to allow risks and uncertainties to be considered. For all intents and purposes, stochastic domains are more challenging when it comes to making decisions, but they are also more flexible than deterministic domains with regard to approximations – in other words, simplifying practical assumptions makes automated decision-making possible in practice. A great number of problem formulations exist, which can be used to represent various aspects and decision-making processes in stochastic domains, with the best-known being decision networks and Markov decision processes.

Many applications require a combination of logical (non-stochastic) and stochastic elements, for example when the control of robots requires high-level specifications in logic and low-level representations for a probabilistic sensor model. Processing natural languages is another area in which this assumption applies, since high-level knowledge in logic needs to be combined with low-level models of text and spoken signals.

## 3.4 Language and communication

In the field of artificial intelligence, processing language is considered to be of fundamental importance, with a distinction being made here between two fields: computational linguistics (CL) and natural language processing (NLP). In short, the difference is that CL research focuses on using computers for language processing purposes, while NLP consists of all applications, including machine translation (MT), Q&A, document summarization, information extraction, to name but a few. In other words, NLP requires a specific task and is not a research discipline per se. NLP comprises:

- Part-of-speech tagging
- Natural language understanding
- Natural language generation
- Automatic summarization
- Named-entity recognition
- Parsing
- Voice recognition
- Sentiment analysis
- Language, topic, and word segmentation
- Co-reference resolution
- Discourse analysis
- Machine translation
- Word sense disambiguation
- Morphological segmentation
- Answers to questions
- Relationship extraction
- Sentence splitting

The core vision of AI says that a version of first-order predicate logic ("first-order predicate calculus" or "FOPC") supported by the necessary mechanisms for the respective problem is sufficient for representing language and knowledge. This thesis says that logic can and should supply the semantics underlying natural language. Although attempts to use a form of logical semantics as the key to representing contents have made progress in the field of AI and linguistics, they have had little success with regard to a program that can translate English into formal logic. To date, the field of psychology has also failed to provide proof that this type of translation into logic corresponds to the way in which people store and manipulate "meaning." Consequently, the ability to translate a language into FOPC continues to be an elusive goal. Without a doubt, there are NLP applications that need to establish logical inferences between sentence representations, but if these are only one part of an application, it is not clear that they have anything to do with the underlying meaning of the corresponding natural language (and consequently with CL/NLP), since the original task for logical structures was inference. These and other considerations have crystallized into three different positions:

- Position 1: Logical inferences are tightly linked to the meaning of sentences, because knowing their meaning is equivalent to deriving inferences and logic is the best way to do this.

- Position 2: A meaning exists outside logic, which postulates a number of semantic markers or primes that are appended to words in order to express their meaning – this is prevalent today in the form of annotations.

- Position 3: In general, the predicates of logic and formal systems only appear to be different from human language, but their terms are in actuality the words as which they appear

The introduction of statistical and AI methods into the field is the latest trend within this context. The general strategy is to learn how language is processed – ideally in the way that humans do this, although this is not a basic prerequisite. In terms of ML, this means learning based on extremely large corpora that have been translated manually by humans. This often means that it is necessary to learn (algorithmically) how annotations are assigned or how part-of-speech categories (the classification of words and punctuation marks in a text into word types) or semantic markers or primes are added to corpora, all based on corpora that have been prepared by humans (and are therefore correct). In the case of supervised learning, and with reference to ML, it is possible to learn potential associations of part-of-speech tags with words that have been annotated by humans in the text, so that the algorithms are also able to annotate new, previously unknown texts.[18] This works the same way for lightly supervised and unsupervised learning, such as when no annotations have been made by humans and the only data presented is a text in a language with texts with identical contents in other languages or when relevant clusters are found in thesaurus data without there being a defined goal.[19] With regard to AI and language, information retrieval (IR) and information extraction (IE) play a major role and correlate very strongly with each other. One of the main tasks of IR is grouping texts based on their content, whereas IE extracts similarly factual elements from texts or is used to be able to answer questions concerning text contents. These fields therefore correlate very strongly with each other, since individual sentences (not only long texts) can also be regarded as documents. These methods are used, for example, in interactions between users and systems, such as when a driver asks the on-board computer a question regarding the owner's manual during a journey – once the language input has been converted into text, the question's semantic content is used as the basis for finding the answer in the manual, and then for extracting the answer and returning it to the driver.

## 3.5 Agents and actions

In traditional AI, people focused primarily on individual, isolated software systems that acted relatively inflexibly to predefined rules. However, new technologies and applications have established a need for artificial entities that are more flexible, adaptive, and autonomous, and that act as social units in multi-agent systems. In traditional AI (see also "physical symbol system hypothesis"[20] that has been embedded into so-called "deliberative" systems), an action theory that establishes how systems make decisions and act is represented logically in individual systems that must execute actions. Based on these rules, the system must prove a theorem – the prerequisite here being that the system must receive a description of the world in which it currently finds itself, the desired target state, and a set of actions, together with the prerequisites for executing these

---

actions and a list of the results for each action. It turned out that the computational complexity involved rendered any system with time limits useless even when dealing with simple problems, which had an enormous impact on symbolic AI, resulting in the development of reactive architectures. These architectures follow if-then rules that translate inputs directly into tasks. Such systems are extremely simple, although they can solve very complex tasks. The problem is that such systems learn procedures rather than declarative knowledge, i.e., they learn attributes that cannot easily be generalized for similar situations. Many attempts have been made to combine deliberative and reactive systems, but it appears that it is necessary to focus either on impractical deliberative systems or on very loosely developed reactive systems – focusing on both is not optimal.

*Principles of the new, agent-centered approach*

The agent-oriented approach is characterized by the following principles:

- Autonomous behavior:

    "Autonomy" describes the ability of systems to make their own decisions and execute tasks on behalf of the system designer. The goal is to allow systems to act autonomously in scenarios where controlling them directly is difficult. Traditional software systems execute methods after these methods have been called, i.e., they have no choice, whereas agents make decisions based on their beliefs, desires, and intentions (BDI)[21].

- Adaptive behavior:

    Since it is impossible to predict all the situations that agents will encounter, these agents must be able to act flexibly. They must be able to learn from and about their environment and adapt accordingly. This task is all the more difficult if not only nature is a source of uncertainty, but the agent is also part of a multi-agent system. Only environments that are not static and self-contained allow for an effective use of BDI agents – for example, reinforcement learning can be used to compensate for a lack of knowledge of the world.

    Within this context, agents are located in an environment that is described by a set of possible states. Every time an agent executes an action, it is "rewarded" with a numerical value that expresses how good or bad the action was. This results in a series of states, actions, and rewards, and the agent is compelled to determine a course of action that entails maximization of the reward.

- Social behavior:

    In an environment where various entities act, it is necessary for agents to recognize their adversaries and form groups if this is required by a common goal. Agent-oriented systems are used for personalizing user interfaces, as middleware, and in competitions such as the RoboCup. In a scenario where there are only self-driving cars on roads, the individual agent's autonomy is not the only indispensable component – car2car communications, i.e., the exchange of information between vehicles and acting as a group on this basis, are just as important. Coordination between the agents results in an optimized flow of traffic, rendering traffic jams and accidents virtually impossible (see also section 5.1, "Vehicles as autonomous, adaptive, and social agents & cities as super-agents").

In summary, this agent-oriented approach is accepted within the AI community as the direction of the future.

*Multi-agent behavior*

Various approaches are being pursued for implementing multi-agent behavior, with the primary difference being in the degree of control that designers have over individual agents.[22,23,24] A distinction is made here between:

- Distributed problem-solving systems (DPS)
- Multi-agent systems (MAS)

DPS systems allow the designer to control each individual agent in the domain, with the solution to the task being distributed among multiple agents. In contrast, MAS systems have multiple designers, each of whom can only influence their own agents with no access to the design of any other agent. In this case, the design of the interaction protocols is extremely important. In DPS systems, agents jointly attempt to achieve a goal or solve a problem, whereas, in MAS systems, each agent is individually motivated and wants to achieve its own goal and maximize its own benefit. The goal of DPS research is to find collaboration strategies for problem-solving, while minimizing the level of communication required for this purpose. Meanwhile, MAS research is looking at coordinated interaction, i.e., how autonomous agents can be brought to find a common basis for communication and undertake consistent actions.[25] Ideally, a world in which only self-driving cars use the road would be a DPS world. However, the current competition between OEMs means that a MAS world will come into being first. In other words, communication and negotiation between agents will take center stage (see also Nash equilibrium).

---

[21] M. Bratman, D. J. Israel, M. E. Pollack: Plans and Resource-Bounded Practical Reasoning, Computational Intelligence, 4: 156-72, 1988

[22] H. Ah. Bond, L. Gasser: Readings in Distributed Artificial Intelligence, San Mateo, CA: Morgan Kaufmann, 1988
[23] E. H. Durfee: Coordination for Distributed Problem Solvers, Boston, MA: Kluwer Academic, 1988
[24] G. Weiss: Multiagent Systems: A Modern Approach to Distributed Artificial Intelligence, Cambridge, MA: MIT Press, 1999

*Multi-agent learning*

Multi-agent learning (MAL) has only relatively recently been bestowed a certain degree of attention.[26,27,28,29] The key problems in this area include determining which techniques should be used and what exactly "multi-agent learning" means. Current ML approaches were developed in order to train individual agents, whereas MAL focuses first and foremost on distributed learning. "Distributed" does not necessarily mean that a neural network is used, in which many identical operations run during training and can accordingly be parallelized, but instead that:

- A problem is split into subproblems and individual agents learn these subproblems in order to solve the main problem using their combined knowledge OR
- Many agents try to solve the same problem independently of each other by competing with each other

Reinforcement learning is one of the approaches being used in this context.[30]

## 4 Data mining and artificial intelligence in the automotive industry

At a high level of abstraction, the value chain in the automotive industry can broadly be described with the following subprocesses:

1. Development
2. Procurement
3. Logistics
4. Production
5. Marketing
6. Sales, after-sales, and retail
7. Connected customer

Each of these areas already features a significant level of complexity, so the following description of data mining and artificial intelligence applications has necessarily been restricted to an overview.

### 4.1 Development

Vehicle development has become a largely virtual process that is now the accepted state of the art for all manufacturers. CAD models and simulations (typically of physical processes, such as mechanics, flow, acoustics, vibration, etc., on the basis of finite element models) are used extensively in all stages of the development process.

The subject of optimization (often with the use of evolution strategies[31] or genetic algorithms and related methods) is usually less well covered, even though it is precisely here in the development process that it can frequently yield impressive results. Multi-disciplinary optimization, in which multiple development disciplines (such as occupant safety and noise, vibration, and harshness (NVH)) are combined and optimized simultaneously, is still rarely used in many cases due to supposedly excessive computation time requirements. However, precisely this approach offers enormous potential when it comes to agreeing more quickly and efficiently across the departments involved on a common design that is optimal in terms of the requirements of multiple departments.

---

[25] K. Frankish, W. M. Ramsey: The Cambridge Handbook of Artificial Intelligence, Cambridge: Cambridge University Press, 2014
[26] E. Alonso: Multi-Agent Learning, Special Issue of Autonomous Agents and Multi-Agent Systems 15(1), 2007
[27] E. Alonso: M. d'Inverno, D. Kudenko, M. Luck, J. Noble: Learning in Multi-Agent Systems, Knowledge Engineering Review 16: 277-84, 2001
[28] M. Veloso, P. Stone: Multiagent Systems: A Survey from a Machine Learning Perspective, Autonomous Robots 8: 345-83, 2000
[29] R. Vohra, M. Wellmann: Foundations of Multi-Agent Learning, Artificial Intelligence 171:363-4, 2007

[30] L. Busoniu, R. Babuska, B. De Schutter: A Comprehensive Survey of Multi-Agent Reinforcement Learning, IEEE Transactions on Systems, Man, and Cybernetics – Part C: Applications and Reviews 38: 156-72, 2008
[31] "Evolution strategies" are a variant of "evolutionary algorithms," which has been developed in Germany. They offer significantly better performance than genetic algorithms for this type of task. See also Th. Bäck: Evolutionary Algorithms in Theory and Practice, Oxford University Press, NY, 1996.

In terms of the analysis and further use of simulation results, data mining is already being used frequently to generate co-called "response surfaces." In this application, data mining methods (the entire spectrum, ranging from linear models to Gaussian processes, support vector machines, and random forests) are used in order to learn a nonlinear regression model as an approximation of the representation of the input vectors for the simulation based on the relevant (numerical) simulation results[32]. Since this model needs to have good interpolation characteristics, cross-validation methods that allow the model's prediction quality for new input vectors to be estimated are typically used for training the algorithms. The goal behind this use of supervised learning methods is frequently to replace computation-time-consuming simulations with a fast approximation model that, for example, represents a specific component and can be used in another application. In addition, this allows time-consuming adjustment processes to be carried out faster and with greater transparency during development.

One example: It is desirable to be able to immediately evaluate the forming feasibility[33] of geometric variations in components during the course of an interdepartmental meeting instead of having to run complex simulations and wait one or two days for the results. A response surface model that has been previously trained using simulations can immediately provide a very good approximation of the risk of excessive thinning or cracks in this type of meeting, which can then be used immediately for evaluating the corresponding geometry.

These applications are frequently focused on or limited to specific development areas, which, among other reasons, is due to the fact that simulation data management, in its role as a central interface between data generation and data usage and analysis, constitutes a bottleneck. This applies especially when simulation data is intended for use across multiple departments, variants, and model series, as is essential for real use of data in the sense of a continuously learning development organization. The current situation in practice is that department-specific simulation data is often organized in the form of file trees in the respective file system within a department, which makes it difficult to access for an evaluation based on machine learning methods. In addition, simulation data may already be very voluminous for an individual simulation (in the range of terabytes for the latest CFD simulations), so efficient storage solutions are urgently required for machine-learning-based analyses.

While simulation and the use of nonlinear regression models limited to individual applications have become the standard, the opportunities offered by *optimizing analytics* are rarely being exploited. Particularly with regard to such important issues as multi-disciplinary (as well as cross-departmental) machine learning, learning based on historical data (in other words, learning from current development projects for future projects), and cross-model learning, there is an enormous and completely untapped potential for increasing efficiency.

## 4.2 Procurement

The procurement process uses a wide variety of data concerning suppliers, purchase prices, discounts, delivery reliability, hourly rates, raw material specifications, and other variables. Consequently, computing KPIs for the purpose of evaluating and ranking suppliers poses no problem whatsoever today. Data mining methods allow the available data to be used, for example, to generate forecasts, to identify important supplier characteristics with the greatest impact on performance criteria, or to predict delivery reliability. In terms of *optimizing analytics*, the specific parameters that an automotive manufacturer can influence in order to achieve optimum conditions are also important.

Overall, the finance business area is a very good field for *optimizing analytics*, because the available data contains information about the company's main success factors. *Continuous monitoring*[34] is worth a brief mention as an example, here with reference to controlling. This monitoring is based on finance and controlling data, which is continuously prepared and reported. This data can also be used in the sense of predictive analytics in order to automatically generate forecasts for the upcoming week or month. In terms of *optimizing analytics*, analyses of the key influencing parameters, together with suggested optimizing actions, can also be added to the aforementioned forecasts.

---

[32] For example: Th. Bäck, C. Foussette, P. Krause: Automatische Metamodellierung von CAE-Simulationsmodellen (Automatic Meta-Modeling of CAE Simulation Models), ATZ – Automobiltechnische Zeitschrift 117(5), 64-69, 2015.

[33] For details: http://www.divis-gmbh.de/fileadmin/download/fallbeispiele/140311_Fallbeispiel_BMW_Machbarkeitsbewertung_Umformsimulation_DE.pdf

[34] This term is used in a great many ways and is actually very general in nature. We use it here in the sense of continuous monitoring of company KPIs that are automatically generated and analyzed on a weekly basis, for example.

These subject areas are more of a vision than a reality at present, but they do convey an idea of what could be possible in the fields of procurement, finance, and controlling.

### 4.3 Logistics

In the field of logistics, a distinction can be made between procurement logistics, production logistics, distribution logistics, and spare parts logistics.

Procurement logistics considers the process chain extending from the purchasing of goods through to shipment of the material to the receiving warehouse. When it comes to the purchasing of goods, a large amount of historical price information is available for data mining purposes, which can be used to generate price forecasts and, in combination with delivery reliability data, to analyze supplier performance. As for shipment, *optimizing analytics* can be used to identify and optimize the key cost factors.

A similar situation applies to production logistics, which deals with planning, controlling, and monitoring internal transportation, handling, and storage processes. Depending on the granularity of the available data, it is possible to identify bottlenecks, optimize stock levels, and minimize the time required, for example here.

Distribution logistics deals with all aspects involved in transporting products to customers, and can refer to both new and used vehicles for OEMs. Since the primary considerations here are the relevant costs and delivery reliability, all the subcomponents of the multimodal supply chain need to be taken into account – from rail to ship and truck transportation through to subaspects such as the optimal combination of individual vehicles on a truck. In terms of used-vehicle logistics, *optimizing analytics* can be used to assign vehicles to individual distribution channels (e.g., auctions, Internet) on the basis of a suitable, vehicle-specific resale value forecast in order to maximize total sale proceeds. GM implemented this approach as long ago as 2003 in combination with a forecast of expected vehicle-specific sales revenue[35].

In spare parts logistics, i.e., the provision of spare parts and their storage, data-driven forecasts of the number of spare parts needing to be held in stock depending on model age and model (or sold volumes) are one important potential application area for data mining, because it can significantly decrease the storage costs.

As the preceding examples show, data analytics and optimization must frequently be coupled with simulations in the field of logistics, because specific aspects of the logistics chain need to be simulated in order to evaluate and optimize scenarios. Another example is the supplier network, which, when understood in greater depth, can be used to identify and avoid critical paths in the logistics chain, if possible. This is particularly important, as the failure of a supplier to make a delivery on the critical path would result in a production stoppage for the automaker. Simulating the supplier network not only allows this type of bottleneck to be identified, but also countermeasures to be optimized. In order to facilitate a simulation that is as detailed and accurate as possible, experience has shown that mapping all subprocesses and interactions between suppliers in detail becomes too complex, as well as nontransparent for the automobile manufacturer, as soon as attempts are made to include Tier 2 and Tier 3 suppliers as well.

This is why data-driven modeling should be considered as an alternative. When this approach is used, a model is learned from the available data about the supplier network (suppliers, products, dates, delivery periods, etc.) and the logistics (stock levels, delivery frequencies, production sequences) by means of data mining methods. The model can then be used as a forecast model in order, for example, to predict the effects of a delivery delay for specific parts on the production process. Furthermore, the use of *optimizing analytics* in this case makes it possible to perform a worst-case analysis, i.e., to identify the parts and suppliers that would bring about production stoppages the fastest if their delivery were to be delayed. This example very clearly shows that optimization, in the sense of scenario analysis, can also be used to determine the worst-case scenario for an automaker (and then to optimize countermeasures in future).

---

[35] http://www.automotive-fleet.com/news/story/2003/02/gm-installs-nutechs-vehicle-distribution-system.aspx

## 4.4 Production

Every sub-step of the production process will benefit from the consistent use of data mining. It is therefore essential for all manufacturing process parameters to be continuously recorded and stored. Since the main goal of optimization is usually to improve quality or reduce the incidence of defects, data concerning the defects that occur and the type of defect is required, and it must be possible to clearly assign this data to the process parameters. This approach can be used to achieve significant improvements, particularly in new types of production process – one example being CFRP[36]. Other potential optimization areas include energy consumption and the throughput of a production process per time unit. *Optimizing analytics* can be applied both offline and online in this context.

When used in offline applications, the analysis identifies variables that have a significant influence on the process. Furthermore, correlations are derived between these influencing variables and their targets (quality, etc.) and, if applicable, actions are also derived from this, which can improve the targets. Frequently, such analyses focus on a specific problem or an urgent issue with the process and can deliver a solution very efficiently – however, they are not geared towards continuous process optimization. Conducting the analyses and interpreting and implementing the results consistently requires manual sub-steps that can be carried out by data scientists or statisticians – usually in consultation with the respective process experts.

In the case of online applications, there is a very significant difference in the fact that the procedure is automated, resulting in completely new challenges for data acquisition and integration, data pre-processing, modeling, and optimization. In these applications, even the provision of process and quality data needs to be automated, as this provides integrated data that can be used as a basis for modeling at any time. This is crucial given that modeling always needs to be performed when changes to the process (including drift) are detected. The resulting forecast models are then used automatically for optimization purposes and are capable of, e.g., forecasting the quality and suggesting (or directly implementing) actions for optimizing the relevant target variable (quality in this case) even further. This implementation of *optimizing analytics*, with automatic modeling and optimization, is technically available, although it is more a vision than a reality for most users today.

The potential applications include forming technology (conventional as well as for new materials), car body manufacture, corrosion protection, painting, drive trains, and final assembly, and can be adapted to all sub-steps. An integrated analysis of all process steps, including an analysis of all potential influencing factors and their impact on overall quality, is also conceivable in future – in this case, it would be necessary to integrate the data from all subprocesses.

## 4.5 Marketing

The focus in marketing is to reach the end customer as efficiently as possible and to convince people either to become customers of the company or to remain customers. The success of marketing activities can be measured in sales figures, whereby it is important to differentiate marketing effects from other effects, such as the general financial situation of customers. Measuring the success of marketing activities can therefore be a complex endeavor, since multivariate influencing factors can be involved.

It would also be ideal if *optimizing analytics* could always be used in marketing, because optimization goals, such as maximizing return business from a marketing activity, maximizing sales figures while minimizing the budget employed, optimizing the marketing mix, and optimizing the order in which things are done, are all vital concerns. Forecast models, such as those for predicting additional sales figures over time as a result of a specific marketing campaign, are only one part of the required data mining results – multi-criteria decision-making support also plays a decisive role in this context.

Two excellent examples of the use of data mining in marketing are the issues of churn (customer turnover) and customer loyalty. In a saturated market, the top priority for automakers is to prevent loss of custom, i.e., to plan and implement optimal countermeasures. This requires information that is as individualized as possible concerning the customer, the customer segment to which the customer belongs, the customer's satisfaction and experience with their current vehicle, and data concerning competitors, their models, and prices. Due to the subjectivity of some of this data (e.g., satisfaction surveys, individual satisfaction values), individualized churn predictions and optimal countermeasures (e.g., personalized discounts, refueling or cash rewards, incentives based on additional features) are a complex subject that is always relevant.

---

[36] C. Sorg: Data Mining als Methode zur Industrialisierung und Qualifizierung neuer Fertigungsprozesse für CFK-Bauteile in automobiler Großserienproduktion (Data Mining as a Method for the Industrialization and Qualification of New Production Processes for CFRP Components in Large-Scale Automotive Production). Dissertation, Technical University of Munich, 2014. Dr. Hut Verlag.

Since maximum data confidentiality is guaranteed and no personal data is recorded – unless the customer gives their explicit consent in order to receive offers as individually tailored as possible – such analyses and optimizations are only possible at the level of customer segments that represent the characteristics of an anonymous customer subset.

Customer loyalty is closely related to this subject, and takes on board the question of how to retain and optimize, i.e., increase the loyalty of existing customers. Likewise, the topic of "upselling," i.e., the idea of offering existing customers a higher-value vehicle as their next one and being successful with this offer, is always associated with this. It is obvious that such issues are complex, as they require information about customer segments, marketing campaigns, and correlated sales successes in order to facilitate analysis. However, this data is mostly not available, difficult to collect systematically, and characterized by varying levels of veracity, i.e., uncertainty in the data.

Similar considerations apply to optimizing the marketing mix, including the issue of trade fair participation. In this case, data needs to be collected over longer periods of time, so that it can be evaluated and conclusions can be drawn. For individual marketing campaigns such as mailing campaigns, evaluating the return business rate with regard to the characteristics of the selected target group is a much more likely objective of a data analysis and corresponding campaign optimization.

In principle, very promising potential applications for *optimizing analytics* can also be found in the marketing field. However, the complexity involved in data collection and data protection, as well as the partial inaccuracy of data collected, means that a long-term approach with careful planning of the data collection strategy is required. The issue becomes even more complex if "soft" factors such as brand image also need to be taken into account in the data mining process – in this case, all data has a certain level of uncertainty, and the corresponding analyses ("What are the most important brand image drivers?" "How can the brand image be improved?") are more suitable for determining trends than drawing quantitative conclusions. Nevertheless, within the scope of optimization, it is possible to determine whether an action will have a positive or negative impact, thereby allowing the direction to be determined, in which actions should go.

## 4.6 Sales, after-sales, and retail

The diversity of potential applications and existing applications in this area is significant. Since the "human factor," embodied by the end customer, plays a crucial role within this context, it is not only necessary to take into account objective data such as sales figures, individual price discounts, and dealer campaigns; subjective customer data such as customer satisfaction analyses based on surveys or third-party market studies covering such subjects as brand image, breakdown rates, brand loyalty, and many others may also be required. At the same time, it is often necessary to procure and integrate a variety of data sources, make them accessible for analysis, and finally analyze them correctly in terms of the potential subjectivity of the evaluations[37] – a process that currently depends to a large extent on the expertise of the data scientists conducting the analysis.

The field of sales itself is closely intermeshed with marketing. After all, the ultimate objective is to measure the success of marketing activities in terms of turnover based on sales figures. A combined analysis of marketing activities (including distribution among individual media, placement frequency, costs of the respective marketing activities, etc.) and sales can be used to optimize market activities in terms of cost and effectiveness, in which case a portfolio-based approach is always used. This means that the optimum selection of a portfolio of marketing activities and their scheduling – and not just focusing on a single marketing activity – is the main priority. Accordingly, the problem here comes from the field of multi-criteria decision-making support, in which decisive breakthroughs have been made in recent years thanks to the use of evolutionary algorithms and new, portfolio-based optimization criteria. However, applications in the automotive industry are still restricted to a very limited scope.

---

[37] One example can be found in this article: http://www.enbis.org/activities/events/current/214_ENBIS_12_in_Ljubljana/programmeitem/1183_Ask_the_Right_Questions__or_Apply_Involved_Statistics__Thoughts_on_the_Analysis_of_Customer_Satisfaction_Data.

Similarly, customer feedback, warranty repairs, and production are potentially intermeshed as well, since customer satisfaction can be used to derive soft factors and warranty repairs can be used to derive hard factors, which can then be coupled with vehicle-specific production data and analyzed. In this way, factors that affect the occurrence of quality defects not present or foreseeable at the factory can be determined. This makes it possible to forecast such quality defects and use *optimizing analytics* to reduce their occurrence. Nevertheless, it is also necessary to combine data from completely different areas – production, warranty, and after-sales – in order to make it accessible to the analysis.

In the case of used vehicles, residual value plays a vital role in a company's fleet or rental car business, as the corresponding volumes of tens of thousands of vehicles are entered into the balance sheet as assets with the corresponding residual value. Today, OEMs typically transfer this risk to banks or leasing companies, although these companies may in turn be part of the OEM's corporate group. Data mining and, above all, predictive analytics can play a decisive role here in the correct evaluation of assets, as shown by an American OEM as long as ten years ago[38]. Nonlinear forecasting models can be used with the company's own sales data to generate individualized, equipment-specific residual value forecasts at the vehicle level, which are much more accurate than the models currently available as a market standard. This also makes it possible to optimize distribution channels – even as far as geographically assigning used vehicles to individual auction sites at the vehicle level – in such a way as to maximize a company's overall sales success on a global basis.

Considering sales operations in greater detail, it is obvious that knowledge regarding each individual customer's interests and preferences when buying a vehicle or, in future, temporarily using available vehicles, is an important factor. The more individualized the knowledge concerning the sociodemographic factors for a customer, their buying behavior, or even their clicking behavior on the OEM's website, as well as their driving behavior and individual use of a vehicle, the more accurately it will be possible to address their needs and provide them with an optimal offer for a vehicle (suitable model with appropriate equipment features) and its financing.

## 4.7 Connected customer

While this term is not yet established as such at present, it does describe a future in which both the customer and their vehicle are fully integrated with state-of-the-art information technology. This aspect is closely linked to marketing and sales issues, such as customer loyalty, personalized user interfaces, vehicle behavior in general, and other visionary aspects (see also section 5). With a connection to the Internet and by using intelligent algorithms, a vehicle can react to spoken commands and search for answers that, for example, can communicate directly with the navigation system and change the destination. Communication between vehicles makes it possible to collect and exchange information on road and traffic conditions, which is much more precise and up-to-date than that which can be obtained via centralized systems. One example is the formation of black ice, which is often very localized and temporary, and which can be detected and communicated in the form of a warning to other vehicles very easily today.

## 5 Vision

Vehicle development already makes use of "modular systems" that allow components to be used across multiple model series. At the same time, development cycles are becoming increasingly shorter. Nevertheless, the field of virtual vehicle development has not yet seen any effective attempts to use machine learning methods in order to facilitate automatic learning that extracts both knowledge that is built upon other historical knowledge and knowledge that applies to more than one model series so as to assist with future development projects and organizing them more efficiently. This topic is tightly intermeshed with that of data management, the complexity of data mining in simulation and optimization data, and the difficulty in defining a suitable representation of knowledge concerning vehicle development aspects. Furthermore, this approach is restricted by the organizational limitations of the vehicle development process, which is often still exclusively oriented towards the model being developed. Moreover, due to the heterogeneity of data (often numerical data, but also images and videos, e.g., from flow fields) and the volume of data (now in the terabyte range for a single simulation), the issue of "data mining in simulation data" is extremely

---

[38] http://www.syntragy.com/doc/q3-05%5B1%5D.pdf

complex and, at best, the object of tentative research approaches at this time[39].

New services are becoming possible due to the use of predictive maintenance. Automatically learned knowledge regarding individual driving behavior – i.e., annual, seasonal, or even monthly mileages, as well as the type of driving – can be used to forecast intervals for required maintenance work (brake pads, filters, oil, etc.) with great accuracy. Drivers can use this information to schedule garage appointments in a timely manner, and the vision of a vehicle that can schedule garage appointments by itself in coordination with the driver's calendar – which is accessible to the on-board computer via appropriate protocols – is currently more realistic than the often cited refrigerator that automatically reorders groceries.

In combination with automatic optimization, the local authorized repair shop, in its role as a central coordination point where individual vehicle service requests arrive via an appropriate telematics interface, can optimally schedule service appointments in real time – keeping workloads as evenly distributed as possible while taking staff availability into account, for example.

With regard to the vehicle's learning and adaptation abilities, there is virtually limitless potential. Vehicles can identify and classify their drivers' driving behavior – i.e., assign them to a specific driver type. Based on this, the vehicles themselves can make adjustments to systems ranging from handling to the electronic user interface – in other words, they can offer drivers individualization and adaptation options that extend far beyond mere equipment features. The learned knowledge about the driver can then be transferred to a new vehicle when one is purchased, ensuring that the driver's familiar environment is immediately available again.

## 5.1 Vision – Vehicles as autonomous, adaptive, and social agents & cities as super-agents

Research into self-driving cars is here to stay in the automotive industry, and the "mobile living room" is no longer an implausible scenario, but is instead finding a more and more positive response. Today, the focus of development is on autonomy, and for good reason: In most parts of the world, self-driving cars are not permitted on roads, and if they are, they are not widespread. This means that the vehicle as an agent cannot communicate with all other vehicles, and that vehicles driven by humans adjust their behavior based on the events in their drivers' field of view. Navigation systems offer support by indicating traffic congestion and suggesting alternative routes. However, we now assume that every vehicle is a fully connected agent, with the two primary goals of:

- Contributing to optimizing the flow of traffic
- Preventing accidents

In this scenario, agents communicate with each other and negotiate routes with the goal of minimizing total travel time (obvious parameters being, for example, the route distance, the possible speed, roadworks, etc.). Unforeseeable events are minimized, although not eliminated completely – for example, storm damage would still result in a road being blocked. This type of information would then need to be communicated immediately to all vehicles in the relevant action area, after which a new optimization cycle would be required. Moreover, predictive maintenance minimizes damage to vehicles. Historical data is analyzed and used to predict when a defect would be highly likely to occur, and the vehicle (the software in the vehicle, i.e., the agent) makes a service appointment without requiring any input from the passenger and then drives itself to the repair shop – all made possible with access to the passenger's calendar. In the event of damage making it impossible to continue a journey, this would also be communicated as quickly as possible – either with a "breakdown" broadcast or to a control center, and a self-driving tow truck would be immediately available to provide assistance, ideally followed by a (likewise self-driving) replacement vehicle. In other words, vehicles act:

- Autonomously in the sense that they automatically follow a route to a destination
- Adaptively in the sense that they can react to unforeseen events, such as road closures and breakdowns
- Socially in the sense that they work together to achieve the common goals of optimizing the flow of traffic and preventing accidents (although the actual situation is naturally more complex and many subgoals need to be defined in order for this to be achieved).

---

[39] L. Gräning, B. Sendhoff: Shape Mining: A Holistic Data Mining Approach to Engineering Design. Advanced Engineering Informatics 28(2), 166-185, 2014.

In combination with taxi services that would also have a self-driving fleet, it would be possible to use travel data and information on the past use of taxi services (provided that the respective user gives their consent) in order to send taxis to potential customers at specific locations without the need for these customers to actively request the taxis. In a simplified form, it would also be possible to implement this in an anonymized manner, for example, by using data to identify locations where taxis are frequently needed (as identified with the use of clusters; see also section 3.1, "Machine learning") at specific times or for specific events (rush hour, soccer matches, etc.).

If roads become digital as well, i.e., if asphalt roads are replaced with glass and supplemented with OLED technology, dynamic changes to traffic management would also be possible. From a materials engineering perspective, this is feasible:

- The surface structure of glass can be developed in such a way as to be skid resistant, even in the rain.

- Glass can be designed to be so flexible and sturdy that it will not break, even when trucks drive over it.

- The waste heat emitted by displays can be used to heat roads and prevent ice from forming during winter.

In this way, cities themselves can be embedded as agents in the multi-agent environment and help achieve the defined goals.

## 5.2 Vision – integrated factory optimization

By using software to analyze customer and repair shop reports and repair data concerning defects occurring in the field, we can already automatically analyze whether an increase in defects can be expected for specific vehicle models or installed parts. The goal here is to identify and avoid potential problems at an early stage, before large-scale recall actions need to be initiated. Causes of defects in the field can be manifold, including deficient quality of the parts being used or errors during production, which, together with the fact that thousands of vehicles leave Volkswagen production plants every day, makes it clear that acting quickly is of utmost importance. Assuming that at present (November 2015), the linguistic analysis of customer statements and repair shop reports shows that a significant increase in right-hand-side parking light failures can be expected for model x, platform C vehicles delivered from July 2015 onwards. In this case, "significant" means that a statistically verifiable upwards trend (increase in reported malfunctions) can be extracted based on vehicle sales between January 2015 and November 2015. By analyzing fault chains and repair chains, it is possible to determine which events will result in a fault or defect or which other models are or will be affected. If the error in production is caused by a production robot, for example, this can be traced back to a hardware fault and/or software error or to an incorrect or incomplete configuration. In the worst-case scenario, it may even be necessary to update the control system in order to eliminate the error. In addition, it is not possible to update software immediately, because work on patches can only begin after the manufacturer has received and reviewed the problem report. Likewise, reconfiguring the robot can be a highly complex task due to the degrees of freedom (axes of rotation) that multi-axis manipulators have at their disposal. In short, making such corrections is time-consuming, demanding, and, in all but ideal scenarios, results in subsequent issues.

Artificial intelligence (AI) approaches can be used to optimize this process at several points.

One of the areas being addressed by AI research is enabling systems (where the term "system" is a synonym for "a whole consisting of multiple individual parts," especially in the case of software or combinations of hardware and software, such as those found in industrial robots) to automatically extract and interpret knowledge from data, although the extent of this is still limited at present.

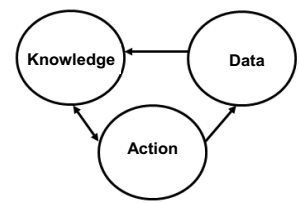

**Figure 4 - Data, knowledge, action**

In contrast to data, knowledge can form the basis for an action, and the result of an action can be fed back into data, which then forms the basis for new knowledge, and so on.

If an agent with the ability to learn and interpret data is supplied with the results (state of the world before the action, state of the world after the action; see also section 3.5) of its own actions or of the actions of other agents, the agent, provided it has a goal and the freedom to adapt as necessary, will attempt to achieve its goal autonomously. Biological agents, such as humans and animals, do this intuitively without needing to actively control or monitor the process of transforming data into knowledge. If, for instance, wood in a DIY project splits because we hammered in a nail too hard at an excessively acute angle, our brain subconsciously transforms the angle, the material's characteristics, and the force of the hammer blow into knowledge and experience, minimizing the likelihood of us repeating the same mistake.

In the previously discussed, specific area of artificial intelligence referred to as "machine learning," research is focused on emulating such behavior. Using ML to enable software to learn from data in a specific problem domain and to infer how to solve new events on the basis of past events opens up a world of new possibilities. ML is nothing new in the field of data analysis, where it has been used for many years now. What is new is the possibility to compute highly complex models with data volumes in the petabyte range within a specific time limit. If one thinks of a production plant as an organism pursuing the objective of producing defect-free vehicles, it is clear that granting this organism access to relevant data would help the organism with its own development and improvement, provided, of course, that this organism has the aforementioned capabilities.

Two stages of development are relevant in this case:

*Stage 1 – Learning from data and applying experiences*

In order to learn from data, a robot must not just operate according to static programming, it must also be able to use ML methods to work autonomously towards defined learning goals. With regard to any production errors that may occur, this means, first and foremost, that the actions being carried out that result in these errors will have been learned, and not programmed based on a flowchart and an event diagram. Assume, for example, that the aforementioned parking light problem has not only been identified, but that its cause can also been traced back to an issue in production, e.g., a robot that is pushing a headlamp into its socket too hard. All that's now required is to define the learning goal for the corrective measure. Let us also assume that the production error is not occurring with robots in other production plants, and that left-hand headlamps are being installed correctly in general.  In the best-case scenario, we, as humans, would be able to visually recognize and interpret the difference between robots that are working correctly and robots that are not – and the robot making the mistake should be able to learn in a similar way. The difference here is in the type of perception involved – digital systems can "see" much better than us in such cases. Even though the internal workings of ML methods implemented by means of software are rarely completely transparent during the learning process – even for the developer of the learning system – due to the stochastic components and complexity involved, the action itself is transparent, i.e., not how a system does something, but what it does. These signals need to be used in order to initiate the learning process anew and to adapt the control system of the problematic robot. In the aforementioned case, these would be the manipulator and effector motion signals of a robot that is working correctly, which can be measured and defined with any desired level of accuracy. This does not require any human intervention, as the system's complete transparency is ensured by continuously securing and analyzing the data accrued in the production process. Neither is any human analysis required in the identification and transmission of defects from the field. Based on linguistic analyses of repair shop and customer reports, together with repair data, we can already swiftly identify which problems are attributable to production. The corresponding delivery of this data to the relevant agents (the data situation makes it possible to determine exactly which machine needs to correct itself) allows these agents to learn from the defects and correct themselves.

*Stage 2 – Overcoming the limitations of programming – smart factories as individuals*

What if the production plant needs to learn things for which even the flexibility of one or more ML methods used by individual agents (such as production or handling robots) is insufficient? Just like a biological organism, a production plant could act as a separate entity composed of subcomponents, similarly to a human, who can be addressed using natural language, understands context, and is capable of interpreting this. The understanding and interpretation of context have always been a challenge in the field of AI research. AI theory views context as a shared (or common) interpretation of a situation, with the context of a situation and the context of an entity relative to a situation being relevant here. Contexts relevant to a production plant include everything that is relevant to production when expressed in natural language or any other way. The following simplified scenario helps in understanding the concept: Let us assume that the final design for a car body is agreed upon by a committee during a meeting.

"We decided on the following body for the Golf 15 facelift. Please build a prototype accordingly based on the Golf 15," says Mr. Müller while looking at the 3-D model that seems to be floating in front of everyone at the meeting and can only be seen with augmented reality glasses.

*In this scenario, use of evolutionary algorithms for simulation is conceivable, limited to the possible combinations that can actually be built. Provided that the required computing power is available and the parameters involved have been reduced, this can cut simulation times from several hours to minutes, making dynamic morphing of components or component combinations possible during a meeting.*

**Factory** : "Based on the model input, I determined that it will take 26 minutes to adjust the programming of my robots. In order to assemble the floor assembly, tools $x_1$, $y_1$ must be replaced with tools $x_2$, $y_2$ on robots x, y. Production of the prototype will be completed in 6 hours and 37 minutes."

Of course, this scenario is greatly simplified, but it should still show what the future may hold. In order to understand what needs to be done, the production plant must understand what a car body is, what a facelift is, etc. and interpret the parameters and output from a simulation in such a way that they can be converted into production steps. This conversion into production steps requires the training of individual ML components on the robots or the adaptation/enhancement of their programs based on the simulation data, so that all steps can be carried out, from cutting sheet metal to assembling and integrating the (still fictitious) Golf 15 basic variant. And while this encompasses a wide range of methods, extending from natural language understanding and language generation through to planning, optimization, and autonomous model generation, it is by no means mere science fiction.

## 5.3 Vision – companies acting autonomously

When planning marketing activities or customer requirements, for example, it is imperative for companies of all types to monitor how sales change over time, to predict how markets will develop and which customers will potentially be lost, to respond to financial crises, and to quickly interpret the potential impact of catastrophes or political structures. We already do all this today, and what we need for it is data. We are not interested in the personal data of individuals, but in what can be derived from many individual components. For example, by analyzing over 1,600 indicators, we can predict how certain financial indicators for markets will move and respond accordingly or we can predict, with a high probability of being correct, which customer groups find models currently in pre-production development appealing and then derive marketing actions accordingly. In fact, we can go so far as to determine fully configured models to suit the tastes of specific customer groups. With this knowledge, we make decisions, adjust our production levels, prepare marketing plans, and propose fully configured models appropriate for specific customer groups in the configurator.

Preparing a marketing plan sometimes follows a static process (what needs to be done), but how something is done remains variable. As soon as it is possible to explain to another person how and why something is being done, this information can also be made available to algorithms. Breaking it down into an example, we can predict that one of our competitors opening a new production plant in a country where we already have manufacturing operations would result in us having to expect a drop in our sales. We can even predict (within a certain fluctuation range) how large this expected drop in sales would be. In this case, "we" refers to the algorithms that we have developed for this specific use case. Since this situation occurs more than once and requires (virtually) identical input parameters every time, we can use the same algorithms to predict events in other countries. This makes it possible to use knowledge from past marketing campaigns in order to conduct future campaigns. In short, algorithms would prepare the marketing plans.

When provided with a goal, such as maximizing the benefit for our customers while taking account of cost-effectiveness, algorithm sub-classes would be able to take internal data (such as sales figures and configurator data) and external data (such as stock market trends, financial indicators, political structures) to autonomously generate output for a "marketing program" or "GDP program." If a company were allowed to use its resources and act autonomously, then it would be able to react autonomously to fluctuations in markets, subsidize vulnerable suppliers, and much more. The possibilities are wide-ranging, and such scenarios can already be technically conceived today. Continuously monitoring stock prices, understanding and interpreting news items, and taking into account demographic changes are only a few of the areas that are relevant and that, among other things, require a combination of natural language understanding, knowledge-based systems, and the ability to make logical inferences. In the field of AI research, language and visual information are very frequently used as the basis for understanding things, because we humans also learn and understand a great deal using language and visual stimuli.

# 6 Conclusions

Artificial intelligence has already found its way into our daily lives, and is no longer solely the subject of science fiction novels. At present, AI is used primarily in the following areas:

- Analytical data processing
- Domains in which qualified decisions need to be made quickly on the basis of a large amount of (often heterogeneous) data
- Monotonous activities that still require constant alertness

In the field of analytical data processing, the next few years will see us transition from exclusive use of decision-support systems to additional use of systems that make decisions on our behalf. Particularly in the field of data analysis, we are currently developing individual analytical solutions for specific problems, although these solutions cannot be used across different contexts – for example, a solution developed to detect anomalies in stock price movements cannot be used to understand the contents of images. This will remain the case in the future, although AI systems will integrate individual interacting components and consequently be able to take care of increasingly complex tasks that are currently reserved exclusively for humans – a clear trend that we can already observe today. A system that not only processes current data regarding stock markets, but that also follows and analyzes the development of political structures based on news texts or videos, extracts sentiments from texts in blogs or social networks, monitors and predicts relevant financial indicators, etc. requires the integration of many different subcomponents – getting these to interact and cooperate is the subject of current research, and new advances in this field are being published every week. In a world where AI systems are able to improve themselves continuously and, for example, manage companies more effectively than humans, what would be left for humans? Time for expanding one's knowledge, improving society, eradicating hunger, eliminating diseases, and spreading our species beyond our own solar system.[40] Some theories say that quantum computers are required in order to develop powerful AI systems[41], and only a very careless person would suggest than an effective quantum computer will be available within the next 10 years. Then again, only a careless person would suggest that it will not. Regardless of this, and as history has taught us time and time again with the majority of relevant scientific accomplishments, caution will also have to be exercised when implementing artificial intelligence – systems capable of making an exponentially larger number of decisions in extremely short times as hardware performance improves can achieve many positive things, but they can also be misused.

---

[40] F. Neukart: Reverse-Engineering the Mind; see https://www.scribd.com/mobile/doc/195264056/Reverse-Engineering-the-Mind, 2014

[41] F. Neukart: On Quantum Computers and Artificial Neural Networks, Signal Processing Research 2(1), 2013